\pdfoutput=1

\documentclass[11pt]{article}
\usepackage{authblk}
\usepackage{graphicx}
\usepackage[]{ACL2023}

\usepackage{times}
\usepackage{latexsym}

\usepackage[T1]{fontenc}

\usepackage[utf8]{inputenc}

\usepackage{microtype}

\usepackage{inconsolata}

\title{Fantastic Semantics and Where to Find Them: Investigating Which Layers of Generative LLMs Reflect Lexical Semantics}

\author[1]{\textbf{Zhu Liu}}
\author[2]{\textbf{Cunliang Kong}}
\author[ \hspace{0.2em}1]{\textbf{Ying Liu}\thanks{\hspace{0.5em} Corresponding author}}
\author[2]{\textbf{Maosong Sun}}

\affil[1]{School of Humanities, Tsinghua University}
\affil[2]{Department of Computer Science and Technology, Tsinghua University}

\affil[ ]{\nolinkurl{{liuzhu22, yingliu,sms}@tsinghua.edu.cn}} 
\affil[ ]{\nolinkurl{cunliang.kong@outlook.com}}

\usepackage{booktabs}
\usepackage{tabularx}
\usepackage{caption}
\usepackage{amsfonts}
\usepackage{amsmath}
\usepackage{color,xcolor} 
\usepackage{xspace}
\usepackage{adjustbox}
\usepackage{multirow}

\newcommand*{\ie}{i.e.\@\xspace}

\usepackage{amssymb}
\usepackage{algpseudocode}
\usepackage{xcolor,colortbl}

\makeatletter

\newcommand{\Rmnum}[1]{\mathrm{\expandafter\@slowromancap\romannumeral #1@}}
\makeatother

\begin{document}

\setcounter{page}{1}

\maketitle

\vspace{2cm} 

\begin{abstract}
Large language models have achieved remarkable success in general language understanding tasks. However, as a family of generative methods with the objective of next token prediction, the semantic evolution with the depth of these models are not fully explored, unlike their predecessors, such as BERT-like architectures. In this paper, we specifically investigate the bottom-up evolution of lexical semantics for a popular LLM, namely Llama2, by probing its hidden states at the end of each layer using a contextualized word identification task. Our experiments show that the representations in lower layers encode lexical semantics, while the higher layers, with weaker semantic induction, are responsible for prediction. This is in contrast to models with discriminative objectives, such as mask language modeling, where the higher layers obtain better lexical semantics. The conclusion is further supported by the monotonic increase in performance via the hidden states for the last meaningless symbols, such as punctuation, in the prompting strategy. Our codes are available at \url{https://github.com/RyanLiut/LLM_LexSem}. 
\end{abstract}

\section{Introduction}
\label{introduction}


GPT-like large language models (LLMs)~\cite{gpt3,touvron2023llama} have recently demonstrated impressive performance on various understanding and generative tasks, shifting from the pretraining-then-finetuning approach employed by BERT-like models~\cite{zhao2023survey}. However, existing research~\cite{ethayarajh2019contextual} suggests that the contextual representations of GPT-like models exhibit subpar performance in downstream tasks, struggling to fully capture the semantic nuances of words. This discrepancy raises a crucial research question: To what extent and through which layers do LLMs encode lexical semantics?

\begin{figure}[t]
    \centering
    \includegraphics[width=0.8\linewidth]{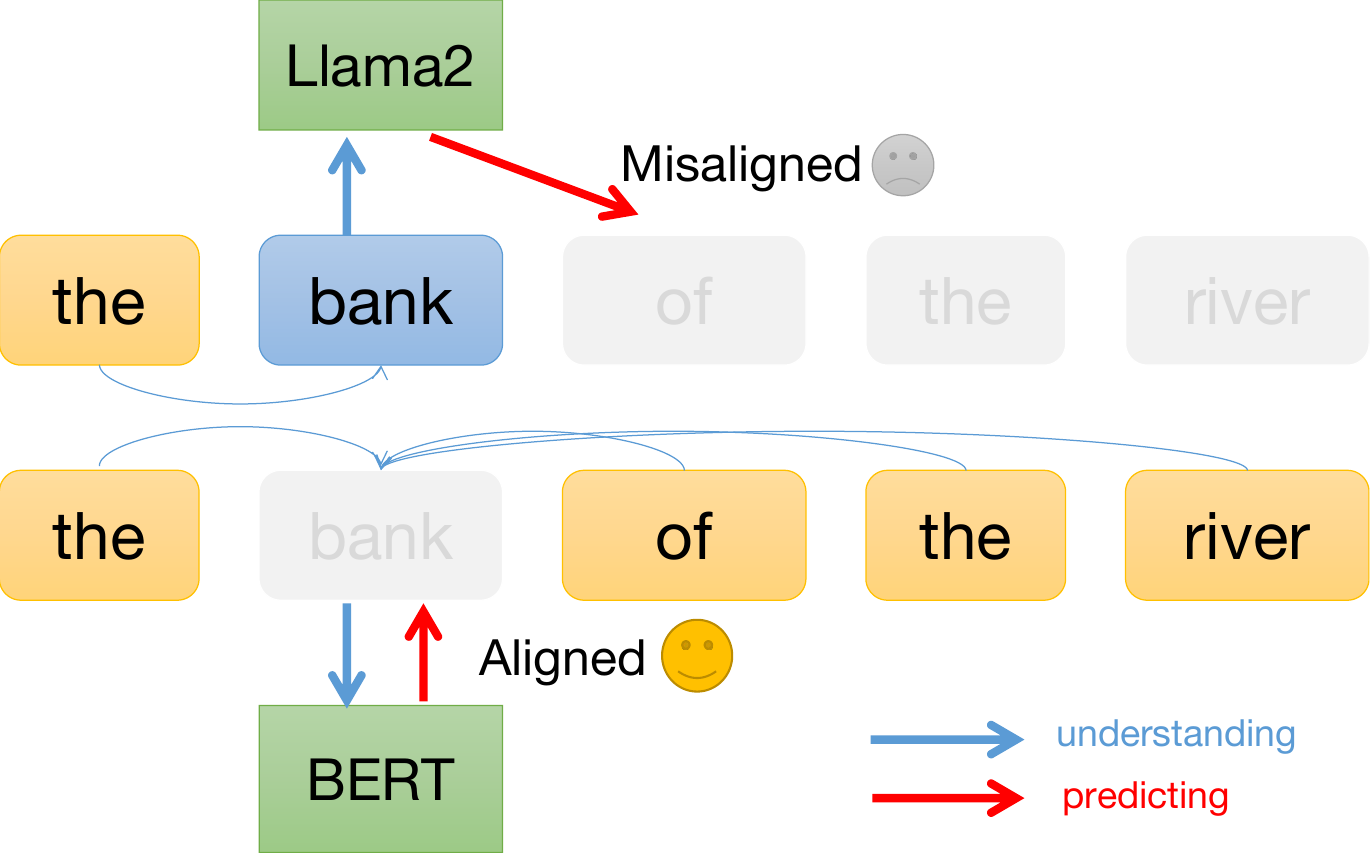}
    \caption{Key Differences between BERT and Llama2 Language Models. Blue and red lines indicate the information flows of understanding and predicting. The "understanding" refers to capture the lexical semantics by leveraging context. The blue line from the context to the current word indicates the flow of understanding. }
    \label{fig:intro}
\end{figure}

Previous research on intermediate layer representations in BERT has revealed important linguistic information, including its hierarchy. For instance, BERT encodes surface features at the bottom, syntactic features in the middle, and semantic features at the top~\cite{jawahar-etal-2019-bert-hierachy}. However, contextual representations in LLMs have received less attention due to structural differences and challenges, as illustrated in Figure~\ref{fig:intro}. Firstly, LLMs employ a decoder-only strategy, which restricts their ability to access only preceding context during inference. Consequently, LLMs struggle to differentiate between homonymous meanings of words such as "bank" in the case of "the bank of the river" and "the bank to save money," due to the shared left context "the". Furthermore, LLMs are trained to predict the next token, resulting in varying degrees of comprehension of historical and predictive contexts across layers~\cite{wang-etal-2023-label,voita-etal-2019-bottom}. In contrast, BERT focuses on masked word restoration through mask language modeling (MLM), where both understanding and prediction processes are targeted for the same word.

In addition, the extraction of contextual word representations from generative Large Language Models (LLMs) often proves to be more beneficial than those derived from models akin to BERT in certain scenarios. Firstly, as generative LLMs are increasingly recognized for their robustness and prevalence as foundational models~\cite{zhao2023survey}, we only obtain ``autoregressive embeddings'' without compromising their generative capabilities. Secondly, these LLMs are consistently trained with billions of parameters across expansive web-scale datasets, thereby exhibiting a significantly greater potential and anticipated superior performance compared to their much lighter counterparts, such as BERT-like models.


Given these observations, we hypothesize that GPT-like LLMs encode lexical semantics in lower layers while making predictions, potentially leading to the forgetting of information related to current tokens in higher layers.\footnote{We refer to lower layers as those closer to the inputs, while higher layers as closer to the outputs.} This hierarchical behavior suggests a dynamic interaction between understanding and prediction in generative LLMs, as indicated by the view of information flow in recent studies~\cite{wang-etal-2023-label,voita-etal-2019-bottom}.

To validate our hypothesis, this study delves into the examination of lexical semantics in LLMs by analyzing how the hidden states at each layer reflect word meanings.
In particular, we investigate the understanding of lexical semantics in the popular open-source LLM, Llama2~\cite{touvron2023llama}, utilizing the word in context benchmark~\cite{pilehvar2019wic}. We employ various input transformation and prompting strategies to fully utilize the contextual information. The results suggest that lower layers of Llama2 capture lexical semantics, while higher layers prioritize prediction tasks. These findings offer practical insights into determining which layers of hidden states to utilize as representations of the meaning of the current word in GPT-like LLMs.

\section{Related Work}
\label{related work}

\subsection{Interpretability of language models}
Interpretability of LLMs can be categorized into mechanistic (bottom-up) and representational (top-down) analysis~\cite{zou2023representation}. Mechanistic interpretability focuses on translating model components into understandable algorithms for humans, typically by representing models as computational graphs and identifying circuits with specific functions~\cite{mecha_olsson2022context,graph_geiger2021causal,function_wang2022interpretability}. On the other hand, representational analysis abstracts away lower-level mechanisms and explores the structure and characteristics of representations. Probing, an effective approach in top-down interpretability, can be classifier-based or geometric-based. Classifier-based probing trains additional classifiers for specific proxy tasks, including syntactic analysis~\cite{hewitt2019syntax}, semantic roles~\cite{semantic_roles_ettinger2020bert}, named entity recognition~\cite{ner2023gpt}, and world knowledge~\cite{world_petroni2019language}. These linguistic features have demonstrated a rich hierarchy, spanning from lower layers to higher layers~\cite{jawahar-etal-2019-bert-hierachy}.
Geometric probing without additional classifiers, examines the properties of the representational space itself. For example, difference vectors, obtained by subtracting base vectors, can detect linguistic features such as scalar adjective intensity~\cite{gari-soler-apidianaki-2021-scalar} and stylistic features~\cite{stylistic_lyu-etal-2023-representation}. 
Furthermore, methods from the view of information flow indicate that models with autoregressive objectives~\cite{voita-etal-2019-bottom} and specifically LLMs~\cite{wang-etal-2023-label} gather information in shallow layers while making predictions in deep layers.

\subsection{Representations of Lexical Semantics}
Lexical semantics, the study of word meanings, is a prominent field in both linguistics and computational research. Linguistics offers rich descriptive entries, known for their high dimensionality, contextual modulation, and discreteness~\cite{break-petersen-potts-2023-lexical}. Early rule-based models, including the Generative Lexicon~\cite{pustejovsky1998generative} approach, used discrete feature representations. In contrast, neural models represent words as compact continuous vectors to avoid arbitrary feature selection. Static vector models, such as word2vec~\cite{word2vec}, Glove~\cite{pennington-etal-2014-glove}, and fastText~\cite{fasttext}, provide unified representations for all word occurrences. To distinguish word meanings in various contexts, especially for polysemous words, researchers have developed context-sensitive representations. Notable models include Elmo~\cite{elmo-peters-etal-2018-deep}, BERT~\cite{kenton2019bert}, and the GPT family~\cite{gpt2,gpt3}. While LSTM-based Elmo and transformer-based models offer bidirectional context around the target word, the GPT family focuses solely on context preceding the query word as a generative model. Large language models (LLMs)~\cite{touvron2023llama, gpt3} follow training mechanisms similar to GPT and have shown competitive performance via prompting engineering~\cite{white2023prompt} compared to BERT-like models, e.g., in lexical tasks like word sense disambiguation~\cite{wsd-kocon2023chatgpt} and named entity recognition~\cite{ner2023gpt}. Our research emphasizes evaluating the quality of representations in LLMs to enhance interpretability, rather than focusing on prompting strategies.

\section{Experimental Design}
\label{sec:experiment}
\subsection{Probing}
We leverage the Word in Context (WiC) dataset as a proxy task for exploring lexical semantics~\cite{pilehvar2019wic}\footnote{\url{https://pilehvar.github.io/wic/}}. This well-structured benchmark presents a binary classification challenge - determining whether identical words convey the same meaning in distinct contexts. Our approach involves utilizing 638 instances from the development set to fine-tune the optimal hyperparameter, and assessing the final performance on 1400 instances from test set. We evaluate results based on accuracy and calculate accuracy separately for instances with different parts of speech.


\subsection{\textcolor{black}{Settings and Models}}
For a given word $w$\footnote{We average the hidden states for tokens within the word as the final word representation.} within context $c$, Llama2 extracts hidden states $h_i \in \mathbb{R} ^ D$ across each of its 32 layers, where $D$ is 4096 in Llama2. The cosine similarity of $w$ in paired contexts ($c^a$, $c^b$) is calculated as $s^{ab}_w$. Subsequently, sentence pairs are classified as true if $s^{ab}_w$ exceeds a threshold $\gamma$, and false if it falls below $\gamma$. The optimal $\gamma$ is determined through development dataset, with distinct values potentially assigned for each layer to accommodate varying similarity ranges. The optimal values of $\gamma$ are listed in Appendix~\ref{sec:threshold}. To address potential anisotropy in the embedding space, we employ standardization across samples following prior research~\cite{ethayarajh2019contextual}.

We employ different input variants for Llama2. The \textbf{base} setting uses the original context $c$ with lexical representations $h_i$ at the target position. Since $w$ cannot access the context behind it in this setting, we repeat the original context and obtain $h_i$ in the second context,  ensuring all information is left of $w$. This configuration is referred to as \textbf{repeat}. We also explore the word before the target one to valid the predictive ability in higher layers, which is denoted as \textbf{repeat\_prev}. Another setting is inspired by the prompting strategy proposed in the paper~\cite{sentence-jiang2023scaling}. Here, we modify the context $c$ as: \textit{The $w$ in this sentence: $c$ means in one word :}. Then, we calculate the representation from the position of the last token, i.e., the final colon \textbf{:}, as $h_i$ and we denote this as \textbf{prompt}. An example is provided in Table~\ref{tab:example_setting}.

\begin{table}
    \small
    \centering
    \begin{tabularx}{\linewidth}{>{\raggedright\arraybackslash}p{0.2\linewidth} p{0.70\linewidth}}
         \toprule
         \multicolumn{1}{c}{setting} & \multicolumn{1}{c}{input} \\
         \midrule
         base & the \textbf{bank} of the river \\
         repeat & the bank of the river the \textbf{bank} of the river \\
         repeat\_prev & the bank of the river \textbf{the} bank of the river \\
         prompt & In this sentence ``the bank of the river'', ``bank'' means in one word \textbf{:}\\
         \bottomrule
    \end{tabularx}
    \caption{An example to show different input formats in three settings. \textbf{Bold} token positions are used as hidden states $h_i$ of target words. We highlight that the bold final colon in the prompt setting is used to extract $h_i$.}
    \label{tab:example_setting}
\end{table}

In order to compare autoregressive generative models with bidirectional models, we conduct experiments on BERT-large\footnote{\url{https://huggingface.co/bert-large-uncased}}, which consists of 25 layers, a hidden dimension of 1024, and 336M parameters. Additionally, we consider other word-level contextualized embedding methods, such as WSD~\cite{wsd-loureiro2019liaad}, Context2vec~\cite{context2vec-melamud-etal-2016-context2vec}, and Elmo~\cite{elmo-peters-etal-2018-deep}, as mentioned in the dataset paper~\cite{pilehvar2019wic}\footnote{It is important to note that we reproduce the result of BERT-large, which is relatively higher than the reported performance in the dataset paper.}.

\section{Results and Analysis}
\label{sec: results}
\begin{table}[t]
    \centering
    \small
    \begin{tabular}{cccc}
    \toprule
        Method & All & Noun & Verb\\
        \midrule
        Human & 80.0 & - & - \\
        Random & 50.0 & - & - \\
        \midrule
        WSD & 67.7 & - & - \\
        BERT\_large\dag (23) & 67.8 & 69.1 & 67.6 \\
        BERT\_large (22) & 71.0 & 70.7 & 71.5 \\
        Context2vec & 59.3 & - & - \\
        Elmo & 57.7 & - & - \\
        \midrule

        Llama2\_base\dag (6) & 60.9 & 63.7 & 58.3 \\
        Llama2\_base (11) & 63.6 & 66.8 & \textcolor{black}{58.7} \\
        Llama2\_repeat\dag (9) & 64.5 & 66.4 & 63.4 \\
        Llama2\_repeat (8) & 68.1 & \underline{72.7} & 65.6 \\
        Llama2\_prompt\dag (28) & \underline{71.1} & 68.9 & \textbf{72.9} \\ 
        Llama2\_prompt (21) & \textbf{72.7} & \textbf{74.5} & \underline{72.1} \\

        \bottomrule
    \end{tabular}
    \caption{Overall accuracy (\%) on the WiC test set. \dag indicates methods without anisotropy removal. The numbers in brackets after the model name indicate the number of layers for achieving the best performance. The index begins at 0, representing the input embedding layer and increases as the model goes deeper. This applies similarly to the remaining indices in the figures.}
    \label{tab:overall}
\end{table}
Table~\ref{tab:overall} presents the overall performance. Llama2, as a generative model, achieves comparable results to bidirectional and non-regressive BERT models, outperforming non-transformer models like Elmo. This suggests that LLMs have the potential for word-level understanding, even though it is not explicitly trained for this capability. As expected, the prompting strategy achieves the highest accuracy among all the Llama2 variants. This approach incorporates downstream tasks into the generative process during LLM training and has proven to be popular and effective in addressing both intermediate and high-level tasks in the LLM era~\cite{zhao2023survey}. However, prompting relies on the choice of prompts and may not directly reveal the model's internal understanding. On the other hand, our repeat strategy demonstrates comparable performance to prompting and significantly outperforms the base version (with a 4.5 advantage gap). This simple yet effective transformation strikes a balance between information accessibility and prompting robustness.

In terms of parts-of-speech, nouns generally exhibit higher accuracy than verbs, as evidenced by a 7.1 advantage gap in Llama2\_repeat. We also observe that in the base setting, verbs exhibit significantly lower accuracy, with decreases of 8.1 points compared to nouns. This decline is attributed to the fact that disambiguating verbs requires more context, which is often lacking in real data preceding the verbs. For instance, target verbs positioned at the beginning of sentences, where there is no prior context to aid in disambiguation, account for 19.2\% of cases, in contrast to 14.3\% for nouns. These observations align with previous studies that have concluded that verbs are more challenging to disambiguate \cite{verb-barba-etal-2021-consec}.

\paragraph{Effectiveness of Anisotropy Removal.} In Table~\ref{tab:overall}, we compare methods with and without anisotropy removal (marked by \dag). The results consistently demonstrate the advantage of methods with anisotropy removal, suggesting that the representation space may collapse into a smaller cone space, as indicated by previous work \cite{ethayarajh2019contextual}. This also offers a simple and practical approach for calculating similarity in the embedding space.

\begin{figure}[h]
    \centering
    \includegraphics[width=1.0\linewidth]{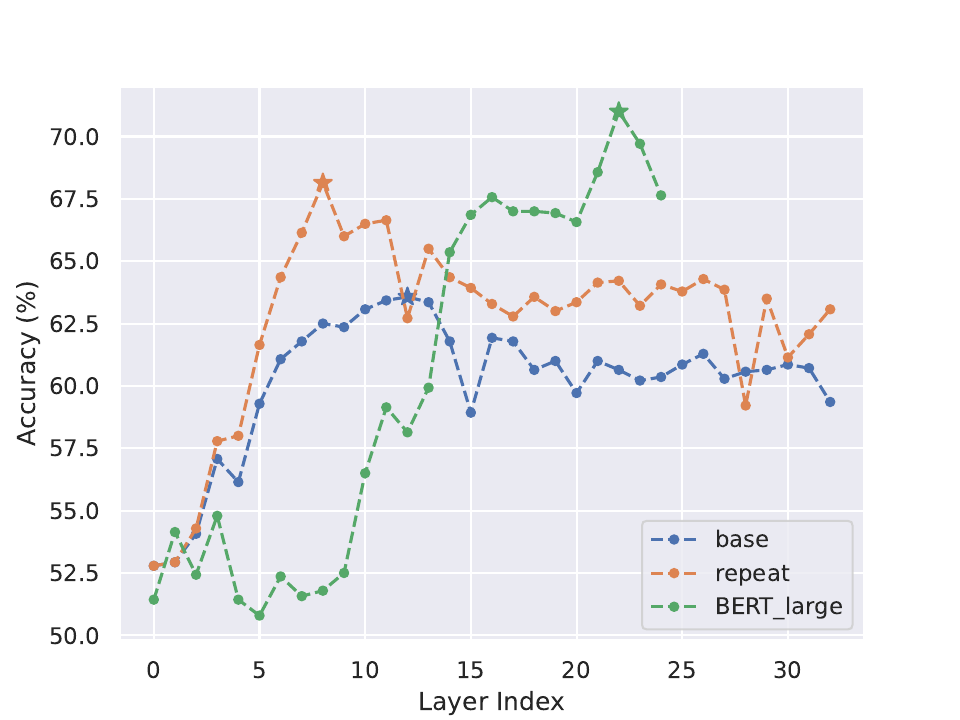}
    \caption{Layer-wise accuracy for different settings and models (Llama2 and BERT\_large). Star shows the best value.}
    \label{fig:layer_wise}
\end{figure}

\paragraph{Trends Across Layers.} Figure~\ref{fig:layer_wise} illustrates the layer-wise dynamics in two settings for Llama2 and also BERT\_large. We observe non-monotonic trends for Llama2 across layers: both the base and repeat initially increase in lower layers before decreasing in higher layers. Consequently, optimal performance is achieved at lower layers when utilizing the hidden states of the target word as the default choice. This suggests that lower layers in LLMs encode lexical semantics, offering both a practical insight and a pathway for interpreting LLMs. Notice that the performance in the highest layer for LLMs does not lose to the worst (\ie, 50\%), indicating it has still remained word meaning in some extent. Moreover, the trend contrasts with bidirectional BERT\_large model, which obtains the best performance in higher layers. This highlights a difference between these two architectures: BERT concentrates on its current word across the layers while Llama2 aims for next token prediction. 


\begin{figure}[h]
    \centering
    \includegraphics[width=1.0\linewidth]{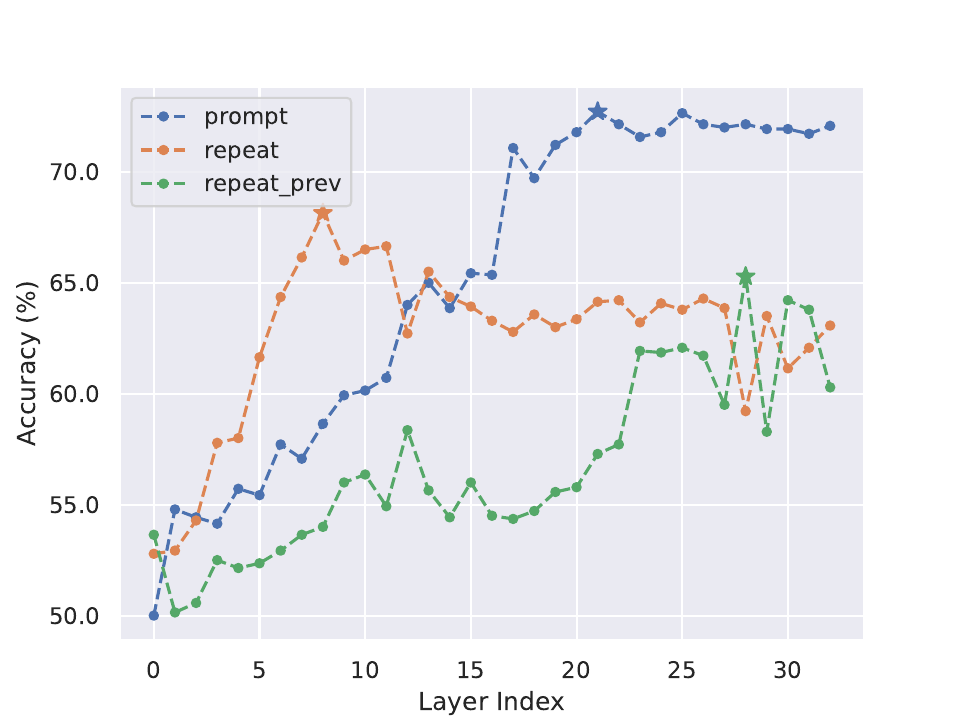}
    \caption{Layer-wise accuracy of Llama2 representations (repeat and prompt setting), as well as the previous token in the repeat setting (repeat\_prev). The increasing trends observed in repeat\_prev and prompt accuracies, as well as the non-monotonic trend observed in repeat accuracy, suggest that while the understanding ability may be weakening, the predictive ability is improving.}
    \label{fig:ntp_comp}
\end{figure}

\paragraph{Balancing Understanding and Prediction.} 
To explore the balance between lexical understanding and predictive capability in Llama2, we computed the accuracy using representations of the previous token before the target (referred to as 
\textbf{repeat\_prev}) in the repeat setting. It is important to note that we opted for the repeat setting instead of the base setting, given that the base setting is constrained by incomplete information access. Furthermore, we conducted a comparison with the prompt setting, as depicted in Figure~\ref{fig:ntp_comp}. Despite the fact that the representations do not originate from the correct target word but are anticipated to represent the next word, both repeat\_prev and prompt exhibit a monotonic trend and comparable result across the layers. This observation suggests that while the understanding may diminish (as indicated by the inverted-U trend in the repeat setting) as layers go deeper, the predictive ability improves.



\section{Conclusion}

This study investigates how Llama2's layer-wise representations encode lexical semantics using the WiC dataset. Our experiments reveal that optimal performance is achieved at lower layers for generative tasks, while predictive accuracy improves in higher layers. This suggests that Llama2 prioritizes understanding before prediction as information flows from lower to higher layers. These findings offer practical guidance on extracting representations for lexical semantics tasks in engineering applications. For example, we would opt to utilize representations from the lower layers for lexical-related tasks, such as POS tagging and word sense disambiguation. Conversely, those from the higher layers could be employed for prediction-related or generative tasks, including text summarization and dialogue generation. Furthermore, it also sheds light on the interpretability of LLMs from a top-down perspective.

\section{Limitations}
Probing offers a valuable viewpoint on lexical semantics, but it is still unclear what kind of semantics representations are exactly learned. Bridging the gap between dense, high-dimensional vectors from computational models and discrete, low-dimensional concepts from linguistic conventions remains an important issue to consider.

Another pressing issue is the narrow focus on only English and one large language model, namely Llama2. Different languages and models may yield varying effects on lexical semantic estimation. We anticipate that future studies will refine and complement our findings using a more diverse sample of natural languages and models.

\section{Ethics Statement}
We do not foresee any immediate negative ethical consequences of our research.

\section{Broader Impact Statement}
Understanding the linguistic knowledge that LLMs have acquired is fundamental for the practical application of generative AI in the real world. This understanding not only enhances the interpretability of these black-box ``Goliaths,'' but also improves the robustness, reliability, and safety of the models. Words carry significant linguistic meaning, while the counterpart tokens serve as the smallest computational units for transformers. We believe that exploring the lexical semantics within LLMs is a foundational step in bridging the gap between computational modeling and linguistics, thereby highlighting the benefits of combining both fields.

\section{Acknowledgements}
The authors thank the anonymous reviewers for their valuable comments and constructive feedback on the manuscript. We also thank members of THUNLP Lab for their valuable discussions. This work is supported by the 2018 National Major Program of Philosophy and Social Science Fund “Analyses and Researches of Classic Texts of Classical Literature Based on Big Data Technology” (18ZDA238) and Research on the Long-Term Goals and Development Plan for National Language and Script Work by 2035 (ZDA145-6).



\bibliography{custom}
\bibliographystyle{acl_natbib}
\clearpage

\appendix

\section{Appendix}
\label{sec:appendix}

\subsection{Optimal Thresholds for each layer}
\label{sec:threshold}
We list the optimal thresholds for each layer in terms of three settings of Llama2, i.e., base, repeat and prompt in Table~\ref{app_tab:thresholds}. They are searched according to the best performance in the development set of WiC dataset.

\begin{table}[h]
    \centering
    \begin{tabular}{cccc}
    \toprule
        Layer Index & base & repeat & prompt \\
        \midrule
0&0.30&0.30&0.00\\
1&0.95&0.95&0.35\\
2&0.90&0.90&0.25\\
3&0.70&0.75&0.35\\
4&0.70&0.70&0.45\\
5&0.40&0.55&0.45\\
6&0.35&0.45&0.45\\
7&0.35&0.40&0.40\\
8&0.30&0.35&0.40\\
9&0.35&0.25&0.45\\
10&0.30&0.25&0.45\\
11&0.30&0.30&0.45\\
12&0.30&0.20&0.50\\
13&0.30&0.30&0.50\\
14&0.30&0.35&0.55\\
15&0.25&0.30&0.55\\
16&0.40&0.35&0.60\\
17&0.40&0.40&0.65\\
18&0.40&0.40&0.60\\
19&0.45&0.40&0.70\\
20&0.45&0.40&0.65\\
21&0.45&0.40&0.65\\
22&0.45&0.40&0.65\\
23&0.40&0.35&0.70\\
24&0.40&0.35&0.65\\
25&0.40&0.35&0.70\\
26&0.40&0.35&0.70\\
27&0.35&0.40&0.70\\
28&0.40&0.20&0.70\\
29&0.40&0.40&0.70\\
30&0.35&0.25&0.70\\
31&0.40&0.25&0.70\\
32&0.35&0.35&0.70\\
\bottomrule
    \end{tabular}
    \caption{Optimal thresholds for each layer.}
    \label{app_tab:thresholds}
\end{table}




\end{document}